\RequirePackage{tikz}

\documentclass[sn-mathphys,Numbered]{sn-jnl}
\usepackage{natbib}

\jyear{2023}%

\theoremstyle{thmstyleone}%
\theoremstyle{thmstyletwo}%

\theoremstyle{thmstylethree}%

\usepackage{pgfplots}
\usepackage{tikz}
\definecolor{bblue}{HTML}{3565c6}
\definecolor{rred}{HTML}{c63535}
\definecolor{ggreen}{HTML}{42c635}
\definecolor{ppurple}{HTML}{a535c6}
\definecolor{yyellow}{HTML}{bfc635}
\definecolor{ccyan}{HTML}{35bdc6}
\definecolor{ppink}{HTML}{c6359a}
\usepackage{subfig}

\usepackage{makecell}

\usepackage{todonotes}

\usepackage{orcidlink}

\begin{document}

\title[Cross-codex Learning for Scribe Identification \textendash PREPRINT]{Cross-codex Learning for Reliable Scribe Identification in Medieval Manuscripts}


\author*[1]{\fnm{Julius} \sur{Weißmann}}\email{weissmann.julius@gmail.com}

\author[1]{\fnm{Markus} \sur{Seidl}}\email{markus.seidl@fhstp.ac.at}

\author[2]{\fnm{Anya} \sur{Dietrich}}\email{a.dietrich@med.uni-frankfurt.de}

\author[3]{\fnm{Martin} \sur{Haltrich}}\email{m.haltrich@stift-klosterneuburg.at}

\affil*[1]{\orgdiv{Media and Digital Technologies}, \orgname{St. Pölten University of Applied Sciences}, \orgaddress{\street{Matthias Corvinus – Straße 15}, \city{St. Pölten}, \postcode{3100}, \state{Lower Austria}, \country{Austria}}}

\affil[2]{\orgdiv{MEG Unit}, \orgname{Brain Imaging Center - Goethe University}, \orgaddress{\street{Heinrich-Hoffmann Strasse 10}, \city{Frankfurt am Main}, \postcode{60528}, \state{Hesse}, \country{Germany}}}

\affil[3]{\orgdiv{Research Center}, \orgname{Klosterneuburg Abbey}, \orgaddress{\street{Stiftspl. 1}, \city{Klosterneuburg}, \postcode{3400}, \state{Lower Austria}, \country{Austria}}}

\abstract{
Historic scribe identification is a substantial task for obtaining information about the past. Uniform script styles, such as the Carolingian minuscule, make it a difficult task for classification to focus on meaningful features. Therefore, we demonstrate in this paper the importance of cross-codex training data for CNN based text-independent off-line scribe identification, to overcome codex dependent overfitting.
We report three main findings:
First, we found that preprocessing with masked grayscale images instead of RGB images clearly increased the F1-score of the classification results.
Second, we trained different neural networks on our complex data, validating time and accuracy differences in order to define the most reliable network architecture.
With AlexNet, the network with the best trade-off between F1-score and time, we achieved for individual classes F1-scores of up to 0,96 on line level and up to 1.0 on page level in classification. 
Third, we could replicate the finding that the CNN output can be further improved by implementing a reject option, giving more stable results.
We present the results on our large scale open source dataset -- the Codex Claustroneoburgensis database (CCl-DB) -- containing a significant number of writings from different scribes in several codices.
We demonstrate for the first time on a dataset with such a variety of codices that paleographic decisions can be reproduced automatically and precisely with CNNs. This gives manifold new and fast possibilities for paleographers to gain insights into unlabeled material, but also to develop further hypotheses.
}

\maketitle

\clearpage

\section{Introduction}\label{secIntro}

In European scriptoria the script \textit{Carolingian minuscule} was used until the second half of 12th century for writing and copying books (codices). 
This Latin script was the first standardized script in the medieval period. Hence, the scribes aimed for a uniform typeface. 
Which scribe wrote which parts of a codex is usually not denoted in the codices. However, the identification of scribes over different codices can help to understand the organization of scriptoria and the travels of codices as well as scribes between the monasteries. The historic auxiliary science paleography deals \textit{inter alia} with the identification of scribes based on certain scribe-typical features of the writing. However, due to the sheer mass of codex pages, this is a tedious and time-consuming task and requires a high level of domain expertise. 

Consequently, only a limited amount of medieval codices has been investigated with a focus on scribe identification. 
The automation utilizing pattern recognition and machine learning allows for larger amounts of material. 
However, to the best of our knowledge data of most automation-based approaches are either limited to even only one \cite{avilaBib, CiliaDaliaStefano, CILIA2020137} or a few different codices or include material from way larger time periods (e.g. \cite{8270156,chammas:hal-03017586}). 

\begin{figure}[t]%
\centering\includegraphics[width=\columnwidth]{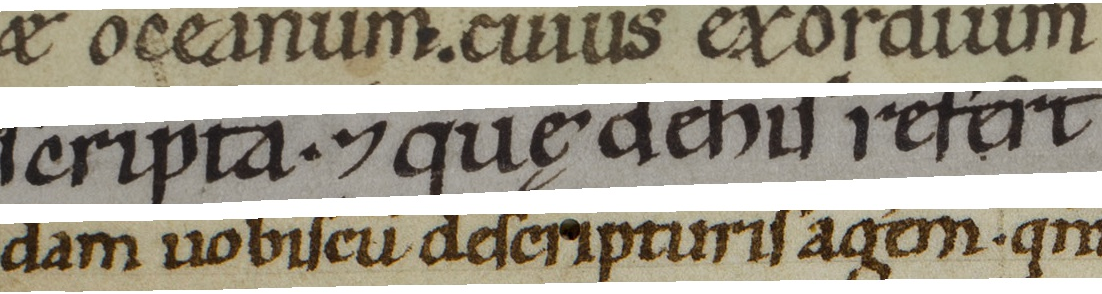}
\caption{Examples from the CCl-DB \cite{ccldbCite} of three lines of different codices from one scribe. The ink and parchment appearance differs, although it's written from the same scribe (class A 30) \citep{haidingerBd2T2,lacknerBd2T3}. Top: CCl 206, middle: CCl 197, bottom: CCl 217}
\label{FigScribeLines}
\end{figure}

To overcome these shortcomings, we investigate automated scribe identification on a large dataset we compiled:  
The CCl-DB \cite{ccldbCite}. 
This dataset contains 51 individual codices with a total amount of 17071 pages. These codices originate from the library of Klosterneuburg Abbey and were written in the late 12th century in Carolingian minuscule \cite{schneider2009palographie,bischoff2004paleoographie}.

We are aiming to answer two central questions: \textit{a) Can the scribe assignments coming from decades of work by paleographic experts} \cite{haidingerBd2T1,haidingerBd2T2,lacknerBd2T3} \textit{ be successfully modelled and predicted ?} and b) \textit{If so, can we use the models to predict scribes for codex pages that have unclear scribe assignments or no scribe assignments at all ?}
A substantial potential data specific risk seen in work by others that could render our work useless is that we could model not only script specific features but also book specific features such as the parchment and page margins. 
To mitigate this risk, we identified scribes that have been found in at least 3 codices. The subset we use contains 25200\footnote{A typical single column page contains 31 or 32 lines. The vast majority of our books is in single column layout, hence we can roughly estimate that the 25200 lines correspond to 800 pages.} random lines uniformly distributed over 7 scribes in 31 different historic codices, in order to train the models to recognize the scribe without codex specific features.

\begin{table}[t]
\begin{center}
\begin{minipage}{\columnwidth}
\caption{Subset of the CCl-DB}\label{TabDataset}
\begin{tabular*}{\columnwidth}{@{\extracolsep{\fill}}lllllll@{\extracolsep{\fill}}}
\toprule%
& \multicolumn{3}{@{}c@{}}{Training codices (\textit{test A} and \textit{B})} & \multicolumn{3}{@{}c@{}}{Separate codices (\textit{test B} only)} \\\cmidrule{2-4}\cmidrule{5-7}%
Class & \raggedleft \vtop{\hbox{\strut Lines}\hbox{\strut per codex}} & \#Codices & Codices & \raggedleft \vtop{\hbox{\strut Lines}\hbox{\strut per codex}} & \#Codices & Codices \\
\midrule
A 30  & 450  & 8 & \raggedleft \vtop{\hbox{\strut 30, 31, 197, 206,}\hbox{\strut 226, 246, 256, 257}}  & 1500 & 2 & 32, 217\\
A 259 & 1200 & 3  & 209, 259, 949  & 1754 & 1 & 706\\
B 259 & 720 & 5  & \raggedleft \vtop{\hbox{\strut 259, 622, 706, 212,}\hbox{\strut 671}} & 1439 & 1 & 246\\
A 20 & 1200 & 3  & 21, 28, 39  & 3000 & 1 & 20\\
B 20 & 900 & 4  & 22, 195, 216, 764  & 119 & 1 & 20\\
A 215 & 1200 & 3  & 215, 219, 703  & 3000 & 1 & 245\\
A 258 & 1800 & 2  & 258, 707  & 3000 & 1 & 203\\
\botrule
\end{tabular*}
\end{minipage}
\end{center}
\footnotetext{The dataset for our experiments is a subset of the seven common scribes of the CCl-DB \cite{ccldbCite}. We generated a dataset, with two groups of codices -- the \textit{training codices} and the \textit{separate codices}. From the \textit{training codices} we choose for every class 3600 random lines, that are uniformly distributed over all books. In total these are 25200 lines that are separated into training, validation and test data (\textit{test A}) according to the ratio of 60 \%, 20 \% and 20 \% respectively. The separate books are used for an extra test set \textit{test B}. There are up to 3000 lines of one class, depending on the codex-size.}
\end{table}

In the last decade, convolutional neural networks (CNNs) have proven to efficiently classify writers in modern and historic context and other tasks such as 
segmentation \cite{DBLP:journals/corr/abs-1804-10371}, optical character recognition \cite{IslamIslamNoor}, and writer identification \cite{XingQiao,avilaBib}. 
For our classification model we compared several general purpose object and concept detection CNNs as well as specific architectures for scribe identification (see \autoref{TabNets}). Surprisingly, the classic AlexNet \cite{KrizhevskySutskeverHintonGeoffrey} 
provided the best trade-off between F1-score and time.
We show that we can distinguish the scribes described by paleographic experts and even identify potential wrong scribe assignments. Furthermore, in combination with the reject option introduced by \cite{CiliaDaliaStefano} we demonstrate that we can reliably predict the scribes for codices with unclear or missing scribe assignments.

In this paper, we focus on three major issues: 
\begin{itemize}
    \item {the importance of cross-codex based training data for automatic scribe identification}
   
    \item {the feasibility of training a model based on scribe assignments by the paleographers Haidinger and Lackner \cite{manuscriptaHaidinger,haidingerBd2T1,haidingerBd2T2,lacknerBd2T3}}
    \item {the necessity of exploiting the confidence in scribe predictions to reveal uncertainties in the dataset.}
\end{itemize}

The latter is a central requirement, as there is no objective ground truth for scribe assignments for the medieval codices we are using.
Our contribution in this paper is threefold:

Firstly, we demonstrate the necessity of book independent training of scribe models, which has been neglected in other studies. 
Secondly, we demonstrate that contrary to the results of Xing and Qiao \cite{XingQiao} standard architectures are sufficiently accurate to reliably identify medieval scribes in a classification pipeline.

Thirdly, our work consequently facilitates comprehensive and convincing studies on large datasets and allows new insights into the historic monastic life and the relationships between the monasteries.

The paper is structured as follows: after an overview about related work in \autoref{secRelWo} we outline the dataset in \autoref{secDataset}. In \autoref{secMethods} we explain the applied methods for scribe identification. We show and discuss the results in \autoref{secResults} and finally conclude the paper in \autoref{secConclusion}.

\begin{figure*}[t]%
\centering
\includegraphics[width=1\textwidth]{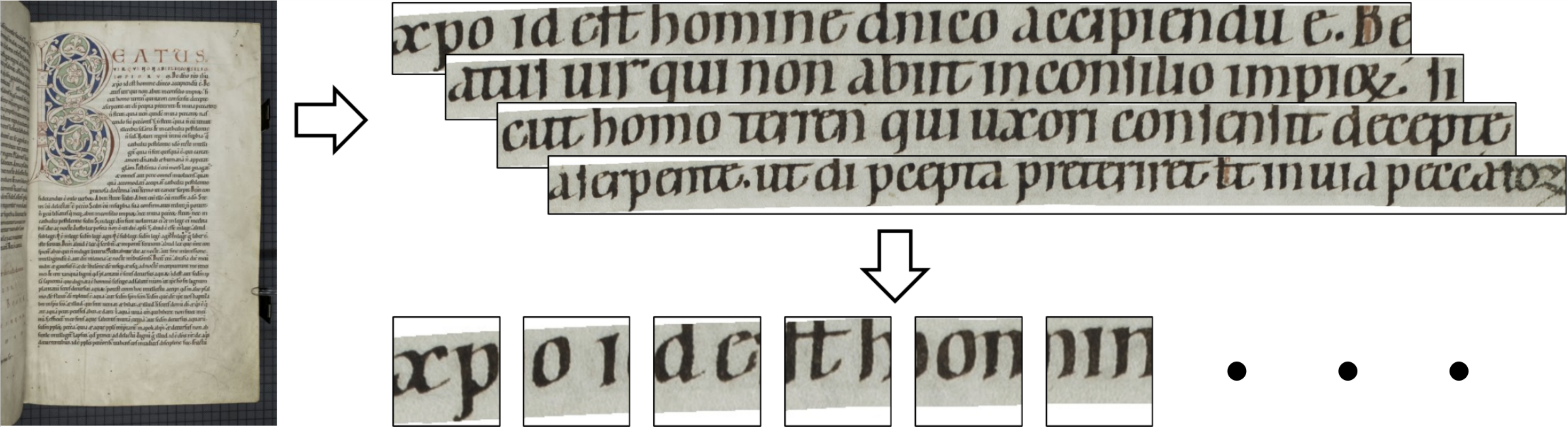}
\caption{Image data of the CCl-DB \cite{ccldbCite} (CCl 20, S. 3r, hand A 20 \cite{haidingerBd2T1}). The CCl-DB provides the codices page by page (left) and on line-level (right-top). The line-level images are produced automatically by the segmentation of Transkribus \cite{8270253}. The neural network classification works with squared images, therefore we cropped the line images into squares and resized them into the network input size}\label{FigCodex}
\end{figure*}

\section{Related work}\label{secRelWo}

Computer-aided historic handwritten document analysis includes segmentation, text recognition, dating and writer identification as well as verification. Segmentation usually separates the written or drawn content from the carrier material (such as parchment or paper)  \cite{DBLP:journals/corr/abs-1804-10371,10.1145/3151509.3151522}. 
Based on this, a possible next step is handwritten text recognition (HTR) \cite{DBLP:journals/corr/abs-1811-07768}. Either the segments or the written content alone or a combination of both modalities allow further investigations like dating, writer verification \cite{DBLP:journals/corr/abs-2009-04532,DBLP:journals/corr/Dey0TGLP17} and identification \cite{chammas:hal-03017586, XingQiao, 8270156}.

Different processes for scribe identification can be used, such as text-dependent and text-independent. Text-dependent \cite{SaidBakerTan} methods allow identifying the writer on particular characters or words, whereas text-independent \cite{YangJinLiu} approaches can be applied on new unseen content. 
Two kinds of handwritten text patterns can be used: on-line and off-line writer identification. On-line \cite{YangJinLiu} systems work with time series of the formation process, while off-line \cite{XingQiao} solutions are limited to the images of the written text document.

Writer identification is a strong topic in document analysis and therefore much discussed. In the last years, a variety of solutions have been provided.  
These methods can be grouped into codebook-based and codebook-free methods.
Codebook-based refer to a codebook that serves as a background model. This model is based on statistical features like in \cite{Maaten05improvingautomatic}. The codebook-free methods include for example the width of ink traces, which was used from Brink et al. \citep{BRINK2012162} to predict the writer in medieval and modern handwritten documents, or the hinge feature provided by Sheng and Lambert \cite{6977065} in order to identify writers in handwritten English text in the IAM-dataset \cite{Marti2002TheIA}. 
Further, there have been strong results in using the Scale-Invariant Feature Transform (SIFT) for writer identification \cite{7333732,FielSablatnig,WuYuan}.

\begin{figure*}[t]%
\centering
\includegraphics[width=1\textwidth]{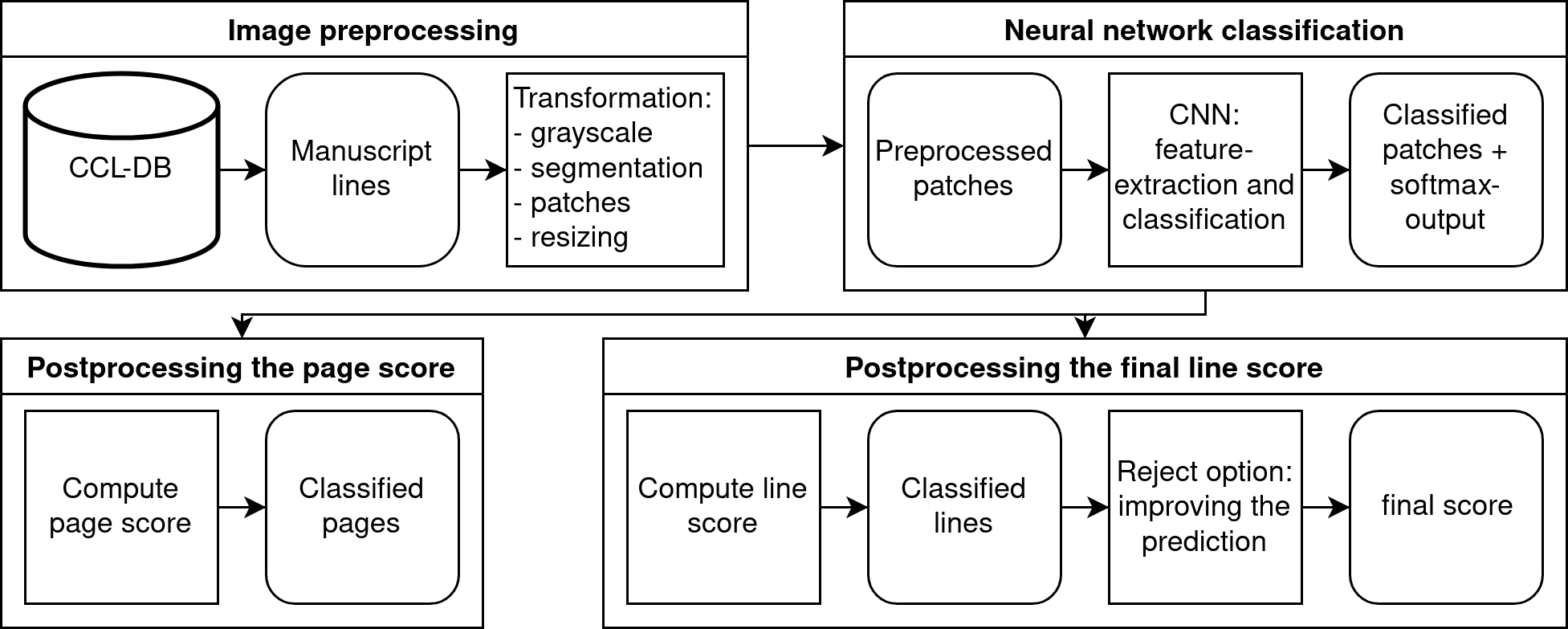}
\caption{Overall procedure of the proposed scribe recognition system. The rounded boxes symbolize data, whereas the angular boxes show processes
}\label{FigPipeline}
\end{figure*}

In recent years, the number of Deep Learning (DL) based studies in document analysis increased drastically  \cite{chammas:hal-03017586, CILIA2020137, XingQiao}. As mentioned in the introduction, the interest in using such techniques is due to their ability to provide powerful state-of-the-art solutions in an efficient and reliable way. During the training, the DL model learns the best fitting features for the classification. Therefore, no handcrafted features are required. 

In \citep{10.1007/978-3-319-23117-4_3} Fiel and Sablatnig presented the strong performance of CNN's for scribe identification on modern data sets.

Xing and Qiao \citep{XingQiao} performed a writer identification on the modern IAM and the HWDB1.1 dataset. They developed a special multi-stream CNN architecture based on AlexNet and outperformed previous approaches on handcrafted features. 

In \cite{CiliaDaliaStefano} Cilia et al. demonstrated a comparison between deep learning and handcrafted features on the Avila Bible that is written in \textit{Carolingian minuscule}.
These classical features, as Cilia et al. call them, are handcrafted features, that have been developed in cooperation with paleographers. The results of their studies emphasize the effectiveness of deep learning features in contrast to the handcrafted features.

\section{Methods for Scribe Identification}\label{secMethods}

In our research, we investigate \textit{scribe} identification in contrast to \textit{writer} identification, since the specific individual of the writing is generally not known for our material.
Scribe identification is discussed in a medieval context only in a very limited range, like at the International Conference on Document Analysis and Recognition (ICDAR) competitions \cite{8270156} or in conjunction with the Avila bible \cite{CiliaDaliaStefano}.
However, the aforementioned datasets are of limited use for our goals. The Avila bible is literally only one codex and consequently does not allow cross-codex scribe identification. The datasets used in the historical ICDAR competitions \cite{8270156,DBLP:journals/corr/abs-1912-03713} span time periods of many centuries, and hence include different scripts and carrier materials. 
Thus, scribe identification in the large amounts of medieval codices in Europe’s libraries is still a challenge and our approach allows novel insights as it focuses on a wide range of codices of several scribes in the short period of the late 12th century.

\begin{table}[t]
\begin{center}
\begin{minipage}{\columnwidth}
\centering
\caption{F1-score for image preprocessing on patch level}\label{TabPreprocess}%
\begin{tabular}{@{}lllll@{}}
\toprule
Data & Network & RGB & GS & GS mask\\
\midrule
Test A & & & & \\
 & AlexNet          & 0.25 & 0.56 & 0.64 \\
 & Deep Writer      & 0.25 & 0.44 & 0.57 \\ 
 & Half Deep Writer & 0.25 & 0.41 & 0.54 \\
 & $\varnothing$    & 0.25 & 0.47 & 0.58 \\
\midrule
Test B & & & & \\
 & AlexNet          & 0.30 & 0.53 & 0.60 \\
 & Deep Writer      & 0.26 & 0.32 & 0.42 \\ 
 & Half Deep Writer & 0.35 & 0.30 & 0.38 \\
 & $\varnothing$    & 0.30 & 0.38 & 0.47 \\
\botrule
\end{tabular}
\end{minipage}
\end{center}
\footnotetext{Preprocessed input images and their impact on the classification of the test data. The experiment was performed on the three different networks AlexNet, Half DeepWriter and Deepwriter. As a result, the averaged F1-score is given. We provide two F1-scores of separate test sets (see \autoref{TabDataset}).}
\end{table}

\section{Dataset}\label{secDataset}

\begin{figure*}[t]%
\centering\includegraphics[width=1\textwidth]{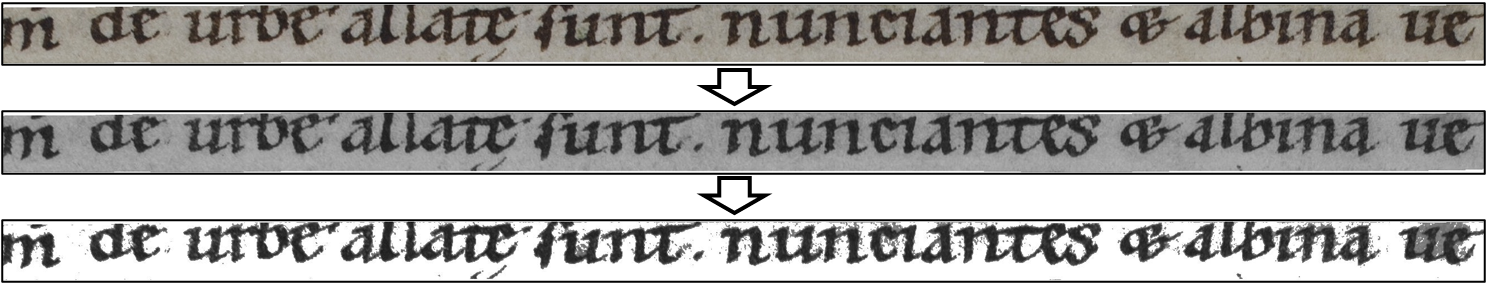}
\caption{Example for the preprocessing (CCl 212, S. 1r, hand B 259 \cite{lacknerBd2T3}). The lines of the CCl-DB \cite{ccldbCite} are provided as RGB images (top). From these, we converted the images to grayscale (middle). The masked grayscaled images are produced by removing the background from the ink}\label{FigPreprocess}
\end{figure*}

We perform experiments on a subset (see \autoref{TabDataset}) of seven  scribes provided by the CCl-DB \citep{ccldbCite}. We selected the scribes which have contributed to as many books as possible to allow cross-codex evaluation.
Samples in the dataset were handwritten on parchment in \textit{Carolingian minuscule} (see \autoref{FigScribeLines} and \ref{FigCodex}) on one- and two-column pages. These codices have been written down in the scriptorium of Klosterneuburg in the last third of the 12th century.
The data is provided by the Scribe ID AI\footnote{See: \url{https://research.fhstp.ac.at/en/projects/scribe-id-ai}} project and has been labelled by paleographic experts based on the activity of the paleographers Haidinger and Lackner \cite{manuscriptaHaidinger,haidingerBd2T1,haidingerBd2T2,lacknerBd2T3}. 52 labelled codices are provided in the CCl-DB. 
This database enables new possibilities within document analysis and especially in handwriting recognition.

To the best of our knowledge, there is no comparable database available that provides the workings of many medieval scribes in various codices and in such a short period of time.

This section will introduce the pipeline of our line-based scribe identification approach. The pipeline is grouped into three main parts (see \autoref{FigPipeline}). In the first part -- the image preprocessing -- we focus on providing the following network input images that are reduced to their scribe specific information.
The second part presents the neural network classification approach. Here we compare different CNNs as image classifiers for scribe identification.
Finally, the third part covers the post-processing which generates the final score. This is where we introduce the computation of the line score and the reject option -- a method to improve the prediction.
All parts will be detailed in the following.

\subsection{Image preprocessing}
The dataset contains not only the codex pages, but also the extracted lines.
The line data is usually correctly extracted from the pages, however there are some small snippets with no noticeable content in the dataset. As image lines are generally of wide aspect ratio, we use a simple heuristic ($ width / height \leq$ 5) to skip these uninformative snippets already in the step of preprocessing.

\begin{table}[t]
\begin{minipage}{\columnwidth}
\centering
\caption{Network performance patch level}\label{TabNets}
\begin{tabular}{@{}llll@{}}
\toprule
Network & F1 Test B & Time & img. (h $*$ w) \\
\midrule
Densenet*       & 0.61 & 503 & 224 $*$ 224 \\
AlexnetNet       & \textbf{0.60} & \textbf{115} & 227 $*$ 227 \\
ResNet18*          & 0.56 & 164 & 224 $*$ 224\\
Inception v3*     & 0.55 & 683 & 299 $*$ 299\\
VGG*              & 0.53 & 610 & 224 $*$ 224\\
SqueezeNet*       & 0.50 & 135 & 224 $*$ 224\\
MNASNet*          & 0.43 & 196 & 224 $*$ 224\\
DeepWriter       & 0.42 & 66  & 113 $*$ 226\\
Half DeepWriter  & 0.38 & 62  & 113 $*$ 113\\
\botrule
\end{tabular}
\end{minipage}
\footnotetext{Nine different CNN architectures are trained on the data of \autoref{TabDataset} to compare their performance on patch level. The weighted averaged F1-Score we use, is measured on \textit{test B} (see \autoref{TabDataset}) and rounded to two decimal places. The training-time is given in rounded minutes. The models marked by * originate from the torchvision library}

\end{table}

For studying the optimal input image data we generated grayscale images and masked grayscale images additionally to the RGB data (see \autoref{FigPreprocess}). 
When we masked the images we followed the example of Stefano et al. \cite{DESTEFANO2018443}. We applied their fixed threshold value, equal to 135, to separate the ink and the parchment of the grayscale images (see \autoref{FigPreprocess}). In related work \cite{CiliaDaliaStefano,8270156,10.1007/978-3-319-23117-4_3,7814129}, binarization is often applied to text images. However, since our masking does not work reliably enough to produce meaningful binarization, we omit this step in this work.

As already mentioned, the lines of the CCl-DB are of different aspect ratio, but the networks we implemented are working with a fixed input size. In order to handle the different lengths of the lines, we followed the patch scanning strategy of Xing and Qiao \cite{XingQiao}. First, the images have been resized in height to the specific network input image height, while maintaining the aspect ratio of the line. Afterwards, we cropped the lines from left to right into patches (see \autoref{FigCodex}). This sliding window comprises the network specific input image width. Due to the large dataset (see \autoref{TabDataset}), there is no need for data augmentation. Hence, we generated patches with no overlap. Only one overlap occurs at the last image of each line, as the last patch of the line is generated by positioning the sliding window at the end of the line.
Finally, we scaled and normalized the patches.

\subsection{Patch level classification}

Xing and Qiao \cite{XingQiao} achieved  high identification accuracies with their customized CNN's on the line level data of the IAM  \cite{Marti2002TheIA} dataset.
They optimized AlexNet to the task of writer identification on the IAM dataset and denoted the architecture Half DeepWriter.
Next, they developed the DeepWriter architecture, which is an improvement of Half DeepWriter that enables the computation of two sequential image patches in a single pass with a multi-stream architecture. 
Xing and Qiao showed that DeepWriter produces the best results on the line level data of the IAM dataset.
Therefore, we implemented these three auspicious architectures as per description of Xing and Qiao \cite{XingQiao}, when we tested the potential of preprocessed images on our data (see \autoref{TabDataset}).

Additionally, we compared several other general-purpose object and concept detection architectures in our study to find the best one suited to our specific data. For this purpose, we used models provided by Torchvision \cite{10.1145/1873951.1874254} (see \autoref{TabNets}) and only adapted the input layer to the grayscale images and the output layer to the seven classes.

As shown in different studies, pre-training and fine-tuning a CNN can lead to better results \cite{XingQiao,Studer2019ACS}. Xing and Qiao \cite{XingQiao} demonstrated this on the IAM \citep{Marti2002TheIA} and the HWDB \cite{LiuYinWangWang} dataset. Therefore, we trained the described models on the IAM dataset, fine-tuned them on our data and compared the results with the from scratch trained weights.  

The models have been trained 
with a batch size of 32 over 10 epochs with a learning rate of $1*10^{-5}$ on ADAM. The error was calculated with the cross entropy. 
Over all ten epochs, the instance of the model that performed best on the validation data has been saved for the next steps of the experiments.


\subsection{Postprocessing}

We pursue a patch, line and page level classification, but as already described, the network classification is on patch level. 
To compute the line and page score, we follow the example of Xing and Qiao \cite{XingQiao}.
They calculate the final score vector $f_i$ for the $jth$ writer, of all patches $N$ of one line $f_i = \frac{1}{N} \sum_{i=1}^N f_{ij}$. 
This averaged Softmax output serves as the basis for the final step of the post-processing -- the reject option.

Cilia et al. \cite{CiliaDaliaStefano} proposed the reject option to generate more reliable results for the writer identification on the Avila bible. 
They showed that sometimes it is better to withdraw a precarious decision than accepting all predictions. In such a case, they reject the prediction with the reject option.  

We used the line score as a probability distribution to check the probabilities for all writers. As Cilia et al. explained, the error-reject curve shows the impact of the reject rate to the wrong predictions and allows finding the optimal threshold for rejecting a prediction. 
Our reject rate is given by the line score of one writer. The reject rate corresponds to the wrong predictions.

\section{Results}\label{secResults}


The purpose of this study was to train a model that classifies reliable and efficiently scribes in cross-codex data. We want to enable the automatic continuation of the work of paleography experts following their example, in order to allow research in large scale. In this study we would like to find a model that is not only reliable but also fast in processing, as it is the basis for research on active learning.

In \autoref{TabPreprocess} we show the importance of cross-codex test data and the risk of overfitting.

In the evaluation of our scribe classification pipeline, we found:
\begin{enumerate} 
\item Image preprocessing plays a key role in cross-codex scribe identification. In comparison to RGB images, masked grayscale images roughly doubled the F1-score in the classification task.
\item Further, we showed that AlexNet provides very fast, and among the most reliable predictions in classifying the scribes of our data set. 
\item Contrary to expectations, pre-training the network on the IAM database leads to worse results, which is why we omitted this step.
\end{enumerate}

Applying the best fitting trained model, it turned out that it is very effective and even indicates incorrect data. 

Moreover, we introduce the reject option on our dataset in order to get rid of precarious classifications and found that it underlines the results.

Finally, we deployed the pipeline to processes open paleographic topics.

\begin{table}[t]
\begin{center}
\begin{minipage}{\columnwidth}
\centering
\caption{Results of \textit{test B}}\label{TabTest}%
\begin{tabular}{@{}lllllll@{}}
\toprule
Scribe & F1-p & F1-l & F1-pg & $\#$-p & $\#$-l & $\#$-pg\\
\midrule
B 259           & 0.60  & 0.76 & 0.85 & 31309  & 1340  & 42  \\
A 259           & 0.79  & 0.94 & 0.98 & 34004  & 1672  & 52  \\
A 30            & 0.62  & 0.76 & 0.74 & 56173  & 2805  & 66  \\
A 20            & 0.90  & 0.96 & 1.00 & 71565  & 2814  & 72  \\
B 20            & 0.05  & 0.24 & 0.00 & 1957   & 111   & 3   \\
A 215           & 0.07  & 0.06 & 0.00 & 65689  & 2897  & 92  \\
A 258           & 0.68  & 0.79 & 0.83 & 67520  & 2893  & 96  \\
\botrule
\end{tabular}
\end{minipage}

\end{center}
\footnotetext{F1-score on \textit{test B} (see \autoref{TabDataset}). The F1-score is measured on patch- (p) and line- (l) and page-level (pg). The weighted averaged test data are random lines of unseen books, as these lines are of various length, they result in a different amount of patches. Due to the image preprocessing the total of samples is slightly lower than in \autoref{TabDataset}.}

\end{table}	

\subsection{Cross-codex data}

The CCl-DB provides handwritings of several scribes in different codices. Therefore, we compared two test sets (see \autoref{TabPreprocess}) to check if the networks tend to learn codex-specific features. For this experiment we trained the architectures AlexNet, Half Deep Writer and DeepWriter. Referring to \cite{XingQiao} these networks are suitable for handwriting identification. We found, that the test set \textit{test B} which contains test samples from books which have not been used for training (see \autoref{TabDataset}) is more comprehensive than \textit{test A} because all three trained networks performed better on \textit{test A} whether the input images have been RGB grayscale or masked. We conclude, that the networks learned codex-specific features in \textit{test A}. Therefore, we used the test set \textit{test B} for further experiments to obtain more reliable results.

\subsection{Classification pipeline}

\begin{figure}[t]%
\centering
\includegraphics[width=1\textwidth]{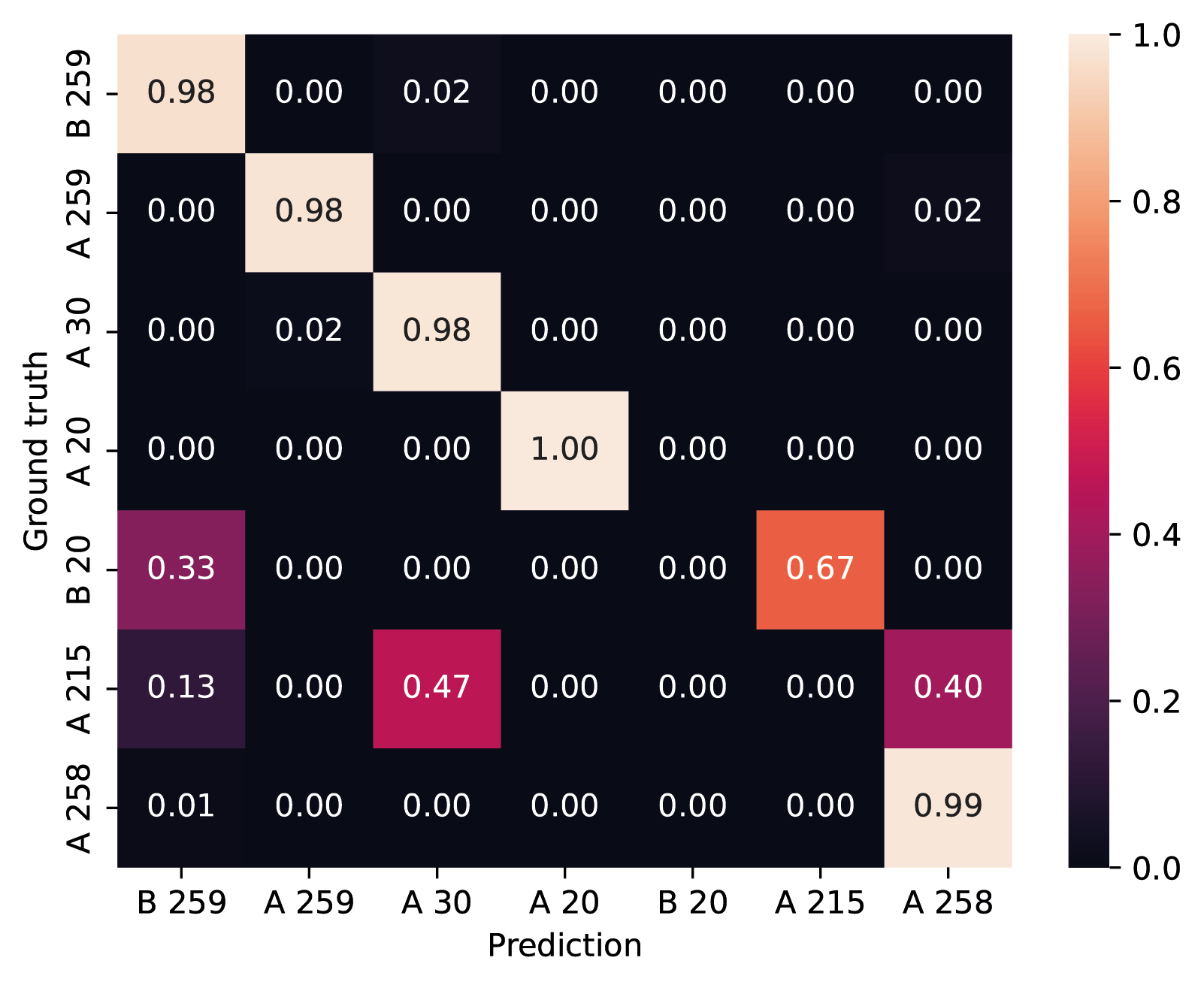}
\caption{Confusion matrix of the data of \textit{test B} on page level (see \autoref{TabNets})}\label{FigConfusion}
\end{figure}

\begin{enumerate} 
	\item To find the best type of input image for the CNN classification, we preprocessed the dataset in three different ways. 
    We compared RGB images with grayscale and masked grayscale images and found that masked grayscale images produce the best F1-score in the test data (see \autoref{TabPreprocess}).
	Consequently, the following experiments are based on this powerful image preprocessing.
	The masking has proven to be effective enough even though we used a simple threshold-based algorithm that sometimes does not reliably distinguish between ink and parchment. Hence, we could replace it in further studies by a better learning-based solution.
	\item As there are different networks available for image classification, we compare in \autoref{TabNets} nine powerful architectures. AlexNet achieves with an F1-score of 0.60 on patch level and 115 minutes training time the best trade-off between F1-score and time. Only DenseNet achieved a small improvement of 0.01 in comparison to AlexNet but with a training time of 503 minutes it is much less time efficient. 
	Even though latest state-of-the-art models performed better, with a growing number of parameters, the processing time would be impractical for our purpose, and as already shown (see \autoref{TabPreprocess}) data-centric approaches such as masked grayscale images are more influential.
	Given these F1 and training time results, we claim AlexNet to be best suited for our purposes and thus used it for all further experiments.
	\item To understand whether pre-training is beneficial, we follow the example of \cite{XingQiao}. Accordingly, we pre-trained AlexNet on 301 writers of the IAM dataset and fine-tuned the model on the CCl-DB data.
	The pre-trained and fine-tuned model generates an F1-score of 0.58 whereas the from scratch trained model outperformed this result with an F1-score of 0.60 on page level.
	Because of the lower F1-score of the pre-trained and fine-tuned model model, we assume that there are not enough shared features between the CCl-DB and the IAM dataset.
	However, as shown by Studer et al. \citep{Studer2019ACS}, models in general benefit from pre-training. We assume, that larger and more comprehensive datasets than the IAM handwriting database could improve our model.
\end{enumerate}
	
\begin{figure}[t]%
\centering
\includegraphics[width=1\textwidth]{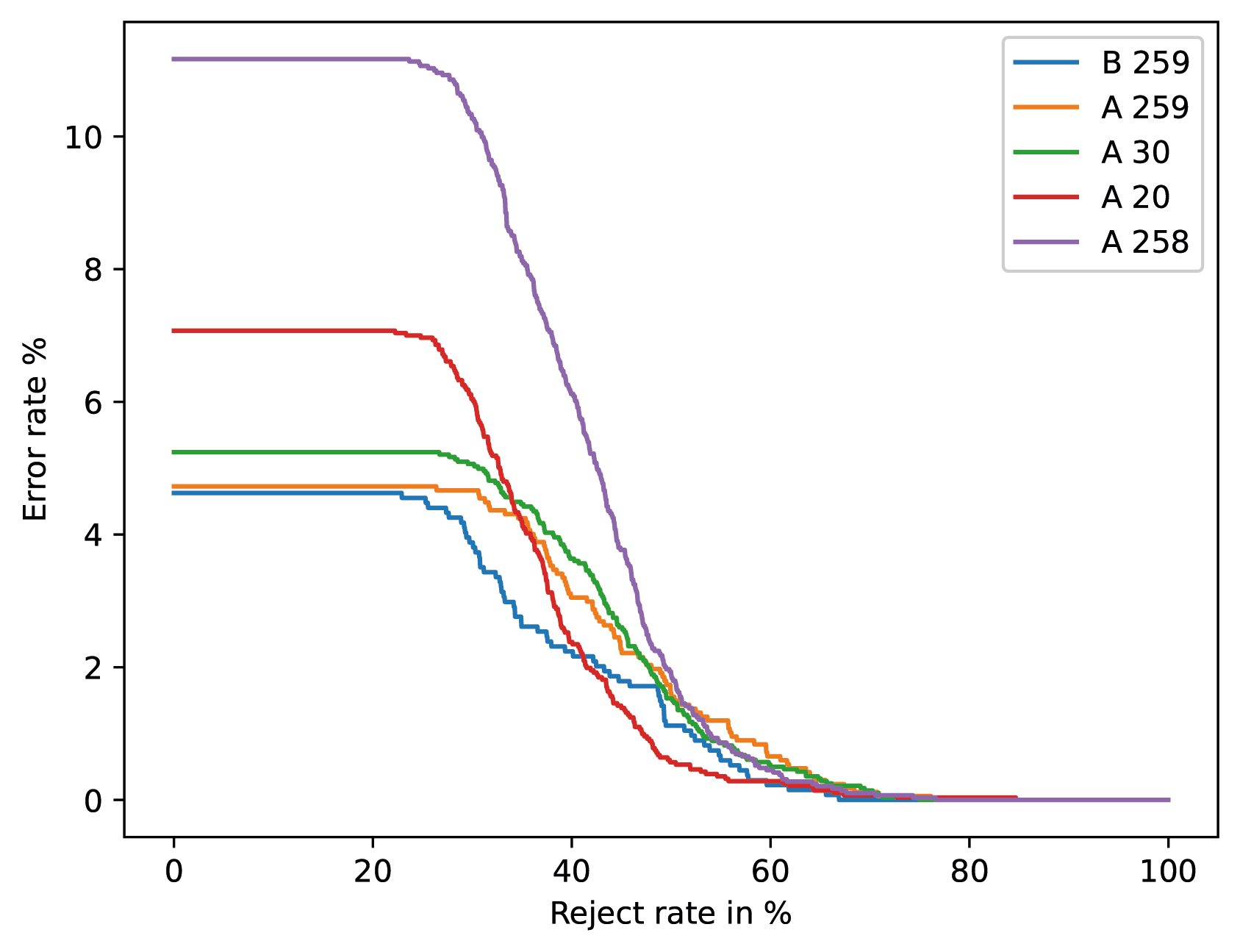}
\caption{Error-reject curves for five different scribes on the data of \textit{test B} on line level (see \autoref{TabDataset})}\label{FigErrorReject}
\end{figure}

\subsection{Automatic paleographic classification}
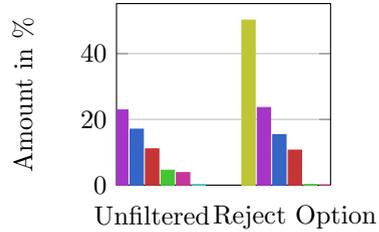
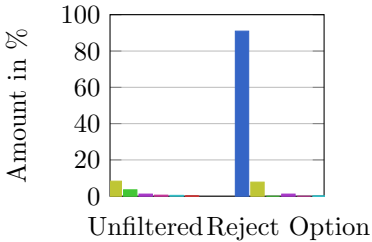
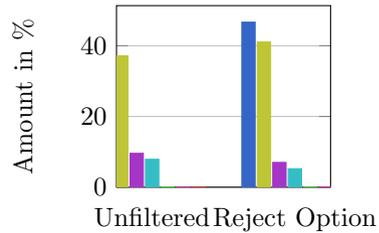
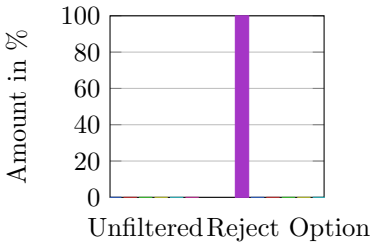
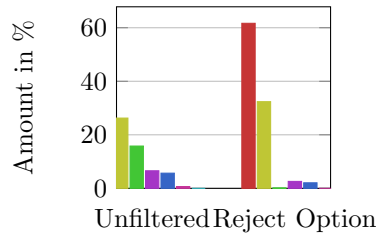
\begin{figure*}[t]%
\centering
\subfloat[\centering Scribe \textit{A30} in CCl \textit{214} on 19110 lines.]{
\begin{tikzpicture}
    \begin{axis}[
        width  = 0.37\textwidth,
        height = 4cm,
        major x tick style = transparent,
        ybar=2*\pgflinewidth,
        bar width=5pt,
        ymajorgrids = true,
        ylabel = {Amount in \%},
        symbolic x coords={Unfiltered,Reject Option},
        xtick = data,
        scaled y ticks = false,
        enlarge x limits=0.25,
        ymin=0,
        legend cell align=left,
        legend style={
                at={(1.05,0)},
                anchor=south west,
                column sep=1ex
        }
    ]
        \addplot[style={bblue,fill=bblue,mark=none}]
            coordinates {(Unfiltered, 63.17) (Reject Option,83.14)};

        \addplot[style={rred,fill=rred,mark=none}]
             coordinates {(Unfiltered,18.88) (Reject Option,2.35)};

        \addplot[style={ggreen,fill=ggreen,mark=none}]
             coordinates {(Unfiltered,13.15) (Reject Option,12.23)};

        \addplot[style={ppurple,fill=ppurple,mark=none}]
             coordinates {(Unfiltered,2.75) (Reject Option,1.37)};
        
        \addplot[style={yyellow,fill=yyellow,mark=none}]
             coordinates {(Unfiltered,0.1) (Reject Option,0.09)};
                     
        \addplot[style={ccyan,fill=ccyan,mark=none}]
             coordinates {(Unfiltered,0.09) (Reject Option,0.0)};
        
        \addplot[style={ppink,fill=ppink,mark=none}]
             coordinates {(Unfiltered,0.09) (Reject Option,0.0)};
        \legend{A30,A20,B20,B259,A259,A258,A215}
    \end{axis}
\end{tikzpicture}}\\
\subfloat[\centering Unknown scribe in CCl \textit{214} on 17 lines]{
\begin{tikzpicture}
    \begin{axis}[
        width  = 0.37\textwidth,
        height = 4cm,
        major x tick style = transparent,
        ybar=2*\pgflinewidth,
        bar width=5pt,
        ymajorgrids = true,
        ylabel = {Amount in \%},
        symbolic x coords={Unfiltered,Reject Option},
        xtick = data,
        scaled y ticks = false,
        enlarge x limits=0.25,
        ymin=0,
        ymax=100,
    ]
        \addplot[style={ccyan,fill=ccyan,mark=none}]
            coordinates {(Unfiltered, 88.24) (Reject Option,100)};

        \addplot[style={ggreen,fill=ggreen,mark=none}]
             coordinates {(Unfiltered,11.76) (Reject Option,0)};

        \addplot[style={rred,fill=rred,mark=none}]
             coordinates {(Unfiltered,0) (Reject Option,0)};

        \addplot[style={ppurple,fill=ppurple,mark=none}]
             coordinates {(Unfiltered,0) (Reject Option,0)};
        
        \addplot[style={yyellow,fill=yyellow,mark=none}]
             coordinates {(Unfiltered,0) (Reject Option,0.09)};
                     
        \addplot[style={bblue,fill=bblue,mark=none}]
             coordinates {(Unfiltered,0) (Reject Option,0.0)};
        
        \addplot[style={ppink,fill=ppink,mark=none}]
             coordinates {(Unfiltered,0) (Reject Option,0.0)};
    \end{axis}
\end{tikzpicture}}
\qquad
\subfloat[\centering First unknown scribe in CCl \textit{213} on 1386 lines]{
\begin{tikzpicture}
    \begin{axis}[
        width  = 0.37\textwidth,
        height = 4cm,
        major x tick style = transparent,
        ybar=2*\pgflinewidth,
        bar width=5pt,
        ymajorgrids = true,
        ylabel = {Amount in \%},
        symbolic x coords={Unfiltered,Reject Option},
        xtick = data,
        scaled y ticks = false,
        enlarge x limits=0.25,
        ymin=0,
    ]
        \addplot[style={yyellow,fill=yyellow,mark=none}]
         coordinates {(Unfiltered,40.62) (Reject Option,50.2)};

        \addplot[style={ppurple,fill=ppurple,mark=none}]
             coordinates {(Unfiltered,22.87) (Reject Option,23.62)};
         
        \addplot[style={bblue,fill=bblue,mark=none}]
            coordinates {(Unfiltered, 17.03) (Reject Option,15.38)};

        \addplot[style={rred,fill=rred,mark=none}]
             coordinates {(Unfiltered,11.04) (Reject Option,10.66)};

        \addplot[style={ggreen,fill=ggreen,mark=none}]
             coordinates {(Unfiltered,04.55) (Reject Option,0.13)};

        \addplot[style={ppink,fill=ppink,mark=none}]
             coordinates {(Unfiltered,03.82) (Reject Option,0.0)};
             
        \addplot[style={ccyan,fill=ccyan,mark=none}]
             coordinates {(Unfiltered,00.07) (Reject Option,0.0)};

    \end{axis}
\end{tikzpicture}}
\qquad
\subfloat[\centering Second unknown scribe in CCl \textit{213} on 1740 lines]{
\begin{tikzpicture}
    \begin{axis}[
        width  = 0.37\textwidth,
        height = 4cm,
        major x tick style = transparent,
        ybar=2*\pgflinewidth,
        bar width=5pt,
        ymajorgrids = true,
        ylabel = {Amount in \%},
        symbolic x coords={Unfiltered,Reject Option},
        xtick = data,
        scaled y ticks = false,
        enlarge x limits=0.25,
        ymin=0,
    ]
        \addplot[style={bblue,fill=bblue,mark=none}]
            coordinates {(Unfiltered, 85.29) (Reject Option,90.97)};
            
        \addplot[style={yyellow,fill=yyellow,mark=none}]
             coordinates {(Unfiltered,8.28) (Reject Option,7.75)};

        \addplot[style={ggreen,fill=ggreen,mark=none}]
             coordinates {(Unfiltered,3.51) (Reject Option,0)};
             
        \addplot[style={ppurple,fill=ppurple,mark=none}]
             coordinates {(Unfiltered,1.15) (Reject Option,1.15)};

        \addplot[style={ppink,fill=ppink,mark=none}]
             coordinates {(Unfiltered,0.57) (Reject Option,0.0)};

        \addplot[style={ccyan,fill=ccyan,mark=none}]
             coordinates {(Unfiltered,0.46) (Reject Option,0.07)};

        \addplot[style={rred,fill=rred,mark=none}]
             coordinates {(Unfiltered,0.11) (Reject Option,0)};

    \end{axis}
\end{tikzpicture}}
\qquad
\subfloat[\centering Third unknown scribe in CCl \textit{213} on 23281 lines]{
\begin{tikzpicture}
    \begin{axis}[
        width  = 0.37\textwidth,
        height = 4cm,
        major x tick style = transparent,
        ybar=2*\pgflinewidth,
        bar width=5pt,
        ymajorgrids = true,
        ylabel = {Amount in \%},
        symbolic x coords={Unfiltered,Reject Option},
        xtick = data,
        scaled y ticks = false,
        enlarge x limits=0.25,
        ymin=0,
    ]
        \addplot[style={bblue,fill=bblue,mark=none}]
            coordinates {(Unfiltered, 43.31) (Reject Option,46.67)};
            
        \addplot[style={yyellow,fill=yyellow,mark=none}]
             coordinates {(Unfiltered,37.15) (Reject Option,41.06)};
             
        \addplot[style={ppurple,fill=ppurple,mark=none}]
             coordinates {(Unfiltered,9.58) (Reject Option,7.01)};
             
        \addplot[style={ccyan,fill=ccyan,mark=none}]
             coordinates {(Unfiltered,7.89) (Reject Option,5.23)};
             
            \addplot[style={ggreen,fill=ggreen,mark=none}]
         coordinates {(Unfiltered,0.06) (Reject Option,0.01)};

        \addplot[style={ppink,fill=ppink,mark=none}]
             coordinates {(Unfiltered,0.06) (Reject Option,0.01)};
             
        \addplot[style={rred,fill=rred,mark=none}]
             coordinates {(Unfiltered,0.01) (Reject Option,0)};

    \end{axis}
\end{tikzpicture}}
\qquad
\subfloat[\centering Fourth unknown scribe in CCl \textit{213} on 25 lines]{
\begin{tikzpicture}
    \begin{axis}[
        width  = 0.37\textwidth,
        height = 4cm,
        major x tick style = transparent,
        ybar=2*\pgflinewidth,
        bar width=5pt,
        ymajorgrids = true,
        ylabel = {Amount in \%},
        symbolic x coords={Unfiltered,Reject Option},
        xtick = data,
        scaled y ticks = false,
        enlarge x limits=0.25,
        ymin=0,
        ymax=100,
    ]
    
        \addplot[style={ppurple,fill=ppurple,mark=none}]
             coordinates {(Unfiltered,100) (Reject Option,100)};
             
        \addplot[style={bblue,fill=bblue,mark=none}]
            coordinates {(Unfiltered, 0) (Reject Option,0)};

        \addplot[style={rred,fill=rred,mark=none}]
             coordinates {(Unfiltered,0) (Reject Option,0)};

        \addplot[style={ggreen,fill=ggreen,mark=none}]
             coordinates {(Unfiltered,0) (Reject Option,0)};

        \addplot[style={yyellow,fill=yyellow,mark=none}]
             coordinates {(Unfiltered,0) (Reject Option,0.0)};
                     
        \addplot[style={ccyan,fill=ccyan,mark=none}]
             coordinates {(Unfiltered,0) (Reject Option,0.0)};
        
        \addplot[style={ppink,fill=ppink,mark=none}]
             coordinates {(Unfiltered,0) (Reject Option,0.0)};
    \end{axis}
\end{tikzpicture}}
\qquad
\subfloat[\centering Fifth unknown scribe in CCl \textit{213} on 899 lines]{
\begin{tikzpicture}
    \begin{axis}[
        width  = 0.37\textwidth,
        height = 4cm,
        major x tick style = transparent,
        ybar=2*\pgflinewidth,
        bar width=5pt,
        ymajorgrids = true,
        ylabel = {Amount in \%},
        symbolic x coords={Unfiltered,Reject Option},
        xtick = data,
        scaled y ticks = false,
        enlarge x limits=0.25,
        ymin=0,
    ]
        \addplot[style={rred,fill=rred,mark=none}]
             coordinates {(Unfiltered,45.05) (Reject Option,61.66)};
             
        \addplot[style={yyellow,fill=yyellow,mark=none}]
             coordinates {(Unfiltered,26.25) (Reject Option,32.38)};
             
        \addplot[style={ggreen,fill=ggreen,mark=none}]
             coordinates {(Unfiltered,15.8) (Reject Option,0.13)};       
             
        \addplot[style={ppurple,fill=ppurple,mark=none}]
             coordinates {(Unfiltered,6.56) (Reject Option,2.59)};
        
        \addplot[style={bblue,fill=bblue,mark=none}]
            coordinates {(Unfiltered, 5.67) (Reject Option,2.07)};

        \addplot[style={ppink,fill=ppink,mark=none}]
             coordinates {(Unfiltered,0.67) (Reject Option,0.0)};             
 
        \addplot[style={ccyan,fill=ccyan,mark=none}]
             coordinates {(Unfiltered,0.0) (Reject Option,0.0)};
    \end{axis}
\end{tikzpicture}}
\caption{Scribe identification on line level and with reject option. All figures are based on the parts of one scribe. These new codices are labeled with one given (a) and six unknown scribes (b-g)}\label{FigScribeId}
\end{figure*}

To figure out, which reliability values can be reached by the trained AlexNet, we evaluated the data of \textit{test B}. The main experiments are performed on line level (see \autoref{FigScribeId}), but we also provide test results on patch and page level (see \autoref{TabTest}). Furthermore, we show the confusion matrix of the same test data on page level in \autoref{FigConfusion}. 
We observe, that the line level classification generally reinforces the patch level results and the page level classification generally reinforces the line level results.

Another particularity in \autoref{TabTest} and \autoref{FigConfusion} are the dichotomous scores.
Five of seven classes are predicted well, such as F1-scores up to 1.0 on page level and 0.96 on line level in the case of A20 (see \autoref{TabTest}).
Only the two classes B 20 and A 215 seem to be less precisely predicted on the test data. We observe low F1-Scores of 0.0 on page level as well as 0.24 and 0.06 on line level respectively.
However, the low predictions for the two classes B 20 and A 215 strengthen the hypothesis of a powerful model as investigations of our paleographic partners revealed the labelling of these classes to be most probably incorrect.
The paleographers actually labeled this test data as several not defined classes. 
Thus, the classification of the two classes proposed by the network might indeed correspond to the correct scribes and could therefore give new insight. However, confirmation by future research using our approach would be needed.

\subsection{Reject Option}

To investigate whether implementing a reject option improves results we tested the five classes B 259, A 259, A 30, A 20, A 258 on it. These are the classes shown previously without conflicts in the test data of \textit{test B}.
\autoref{FigErrorReject} shows that increasing the reject rate minimizes the error.
The reject curves of \autoref{FigErrorReject}  drop quickly, indicating a strong influence of the threshold. 
As Cilia et al. \citep{CiliaDaliaStefano} explained, the reject option is therefore suitable for the scribe identification on the CCl-DB.
Therefore, we choose a reject option based on the threshold of 40 \%, as it ensures low error with high sample rate.
As class B 20 and A 215 are difficult to evaluate from the test data, the threshold is adapted to 60 \%.

\subsection{Into the wild: Using our model on data with unknown scribes}

The central aim of our approach is to contribute new insights for paleography. Therefore, we examined sections from the codices that the experts limited to one scribe, although they could not determine the exact individual. 
As shown in \autoref{FigScribeId} the trained model contributes meaningful classifications for these parts. 
The first plot can be considered as reference, this is the part of CCl 214 written by A 30. In this example, the model recognizes class A 30 as main class.
The remaining six plots can be differentiated into two groups. On the one hand there are plots b, d and f which give significant results that refer to one scribe class B 20, A 30 and A 215 respectively. On the other hand, there are plots like in c, e and g that produce diffuse classification, not focused on one class. We conclude that these less significant predictions are caused either by scribes the model hasn't learned yet or by more than one scribe.

\section{Conclusion}\label{secConclusion}

In this paper, we want to study the question of how to train a reliable and efficient model that allows cross-codex scribe identification in the strongly standardized medieval Carolingian minuscule of the CCl-DB.
To this aim, we first figured out the risk of codex specific overfitting and showed the importance of cross-codex data to overcome this issue. 
We also found, that the reduction of RGB-images to grayscale masked images helps the network to focus on scribe specific features and leads to significantly better results.

After comparing several networks, AlexNet was used in our pipeline to generate a classification on patch, line and page level. Finally, we improved the final score by implementing the reject option. 

One of the limitations of the proposed method is the basic segmentation, which is challenging on the historic parchment.
This limitation leads to a natural direction of future work, focusing on improving the segmentation method that also allows binarization. In a broader context we see future work to use our approach - which allows efficient scribe predictions for unseen books - in an active learning loop that leverages the expert knowledge from paleographers. A visual interface could allow expert verification and correction of the predictions with the goal to iteratively re-train and test our model with new scribe hypotheses.

\section*{Funding}\label{Funding}
This work has received funding from the Lower Austrian FTI-Call 2018 under grant agreement No FTI18-004 (project Active Machine Learning for Automatic Scribe Identification in 12th Century Manuscripts). Moreover, the work was supported by ERASMUS+ from the German Academic Exchange Service (DAAD).

\section*{Acknowledgments}\label{Acknowledgments}

We would like to thank the team of the Klosterneuburg abbey library, as well as the team of the institute of Creative\textbackslash Media/Technologies of the St. P\"olten University of Applied Sciences for their help.


%
\end{document}